\documentclass{llncs}

\usepackage{amsmath}
\usepackage{multirow}
\usepackage{hhline}
\usepackage{graphicx}
\usepackage{caption}
\usepackage{subcaption}
\usepackage{wrapfig}
\usepackage[table,xcdraw]{xcolor}
\usepackage{pgfplots}
\begin{document}
\setlength{\abovedisplayskip}{5pt}
\setlength{\belowdisplayskip}{2pt}
\title{Generalizability \textit{vs.} Robustness: Exploring Adversarial Examples in Medical Imaging }
\title{Generalizability \textit{vs.} Robustness: \\ Adversarial Examples for Medical Imaging }

%
\titlerunning{Generalizability \textit{vs.} Robustness of medical imaging networks}  
%
 \author{Magdalini Paschali\inst{1} \and Sailesh Conjeti\inst{2} \and Fernando Navarro\inst{1}\and Nassir Navab\inst{1,3}}
\authorrunning{Magdalini Paschali et al.} 
%
\tocauthor{Magdalini Paschali, Sailesh Conjeti, Fernando Navarro, Nassir Navab}
%
\institute{
Computer Aided Medical Procedures, Technische Universit\"{a}t M\"{u}nchen, Germany
\and
German Center for Neurodegenerative Diseases (DZNE), Bonn, Germany
\and 
Computer Aided Medical Procedures, Johns Hopkins University, USA}



\maketitle              

\begin{abstract}
In this paper, for the first time, we propose an evaluation method for deep learning models that assesses the performance of a model not only in an unseen test scenario, but also in extreme cases of noise, outliers and ambiguous input data. To this end, we utilize adversarial examples, images that fool machine learning models, while looking imperceptibly different from original data, as a measure to evaluate the robustness of a variety of medical imaging models. Through extensive experiments on skin lesion classification and whole brain segmentation with state-of-the-art networks such as Inception and UNet, we show that models that achieve comparable performance regarding generalizability may have significant variations in their perception of the underlying data manifold, leading to an extensive performance gap in their robustness.
\end{abstract}


\section{Introduction}

Deep learning is being increasingly adopted within the medical imaging community for a plethora of tasks including classification, segmentation, detection \textit{etc.} The classic approach towards the assessment of any machine learning model revolves around the evaluation of its \textit{generalizability} \textit{i.e.} its performance on unseen test scenarios. 
However, in case of \textit{limited} training data, such as medical imaging datasets, using heavily over-parameterized deep learning models could lead to the "memorization" of the training data. Evaluating such models on an available non-overlapping test set is popular, yet significantly limited in its ability to explore the model's resilience to outliers and noisy data / labels (\textit{i.e.} robustness). Additionally, the limited interpretability of deep learning models due to their "black-box" nature challenges their adoption into clinical practice. 

Existing model evaluation routines look deeply into over-fitting but insufficiently into scenarios of model sensitivity to variations of the input. 
Robustness evaluation estimates potential failure probabilities when the model is \textit{pushed} to its limits. In this paper, we approach evaluating a model by leveraging adversarial examples~\cite{szegedy} that are crafted with the purpose of \textit{fooling} a model and can uncover cases where its performance may degenerate. Our approach to using adversarial examples as benchmark is also significantly less laborious and expensive than constituting a sufficiently diverse test set with manual annotation.



\setlength{\columnsep}{0.5pt}%
\begin{figure}
\centering
  \includegraphics[width=0.8\linewidth]{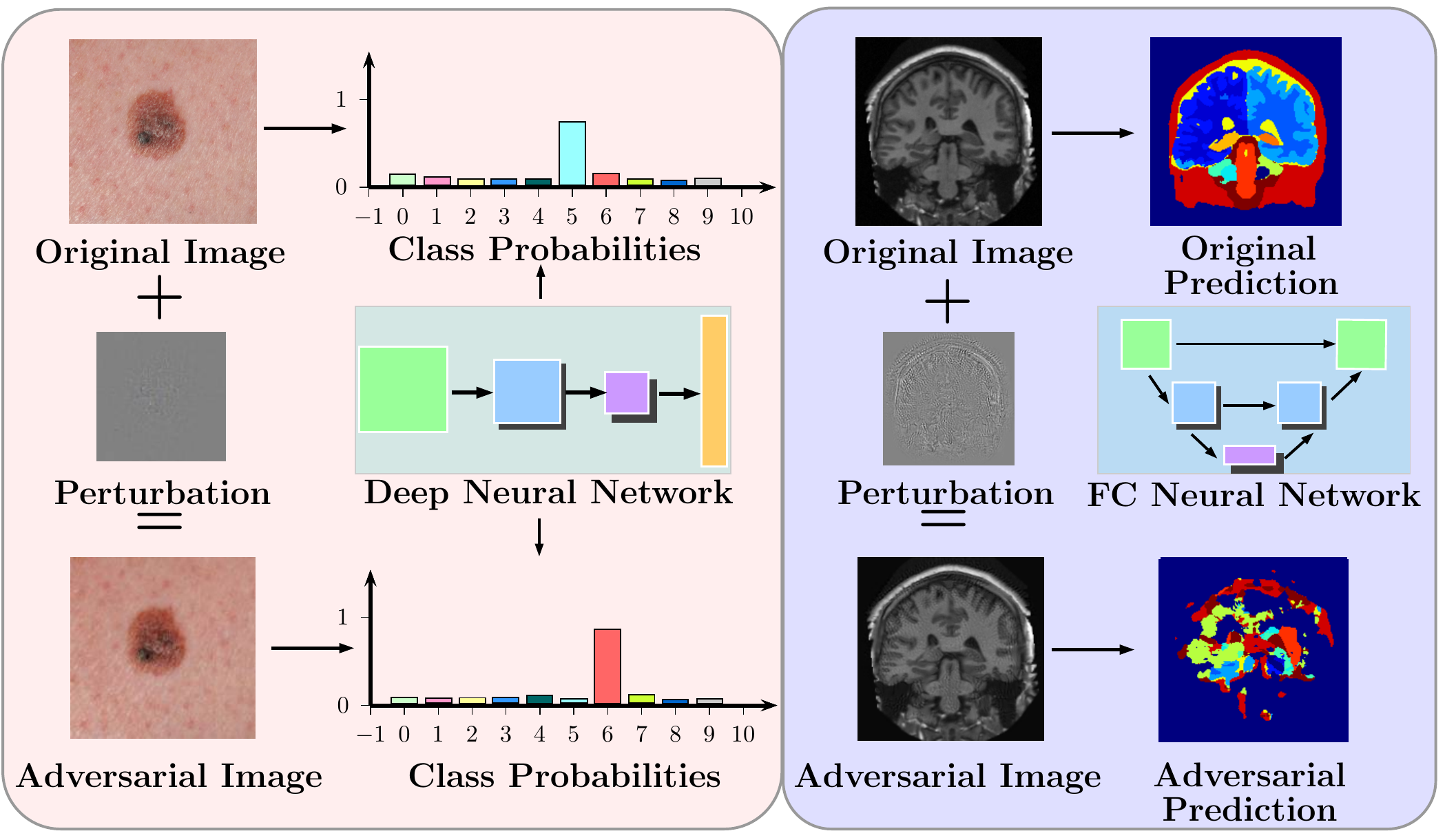}
  \captionof{figure}{\small{Overview of Adversarial Crafting and its effect on network prediction. The difference between the generated adversarial image and the original image is imperceptible, yet deep neural networks are successfully fooled into anomalous predictions.}}
  \label{fig:intro}
\end{figure} 
\noindent

Adversarial examples are images crafted to purposely fool machine learning models, while the added perturbations are imperceptible to human eyes~\cite{szegedy}, as shown in  Fig.~\ref{fig:intro}. To the best of our knowledge, this is the first paper to explore adversarial examples in medical image computing and leverage them in a constructive fashion to benchmark model performance not only on clean and noisy but also on adversarially crafted data. It must be noted that though these examples may not occur in naturally acquired data, utilizing them can present new opportunities for medical imaging researchers to uncover more about their models, with the ultimate goal of increasing robustness and optimizing the decision boundaries learned for different tasks. 

Our contribution is two-fold: Firstly, we demonstrate on a variety of medical image computing tasks that widely adopted state-of-the art deep learning models are not immune to adversarial examples crafting. Secondly, we utilize adversarial examples to benchmark model robustness by comparing a variety of architectures, such as Inception~\cite{inception} and UNet~\cite{unet}, for the tasks of skin lesion classification and whole brain segmentation.

\section{Methodology}
\subsection{Adversarial Crafting}
\label{sec:advCraft}
Given a trained model $F$, an original input image $X$ with output label $Y$, we generate an adversarial example $\hat{X}$ by solving a box-constrained optimization problem $\text{min}_{\hat{X}}  \| \hat{X} - X  \|$ subject to $F\left ( X \right ) = Y$, $F ( \hat{X}  ) = \hat{Y}$, $\hat{Y} \neq Y$ and $\hat{X} \in \left [ 0,1 \right ]$.  
Such an optimization minimizes the added perturbation, say $r$ (\textit{i.e.} $\hat{X} = X + r$) while simultaneously \textit{fooling} the model $F$~\cite{szegedy}. By imposing an additional constraint such as $\left \| r \right \| \leq \epsilon $, we can restrict the perturbation to be small enough to be imperceptible to humans.




\begin{figure}[t]
\centering
\begin{minipage}{\textwidth}
  \begin{minipage}[t]{0.55\textwidth}
\centering
\includegraphics[width=0.975\textwidth]{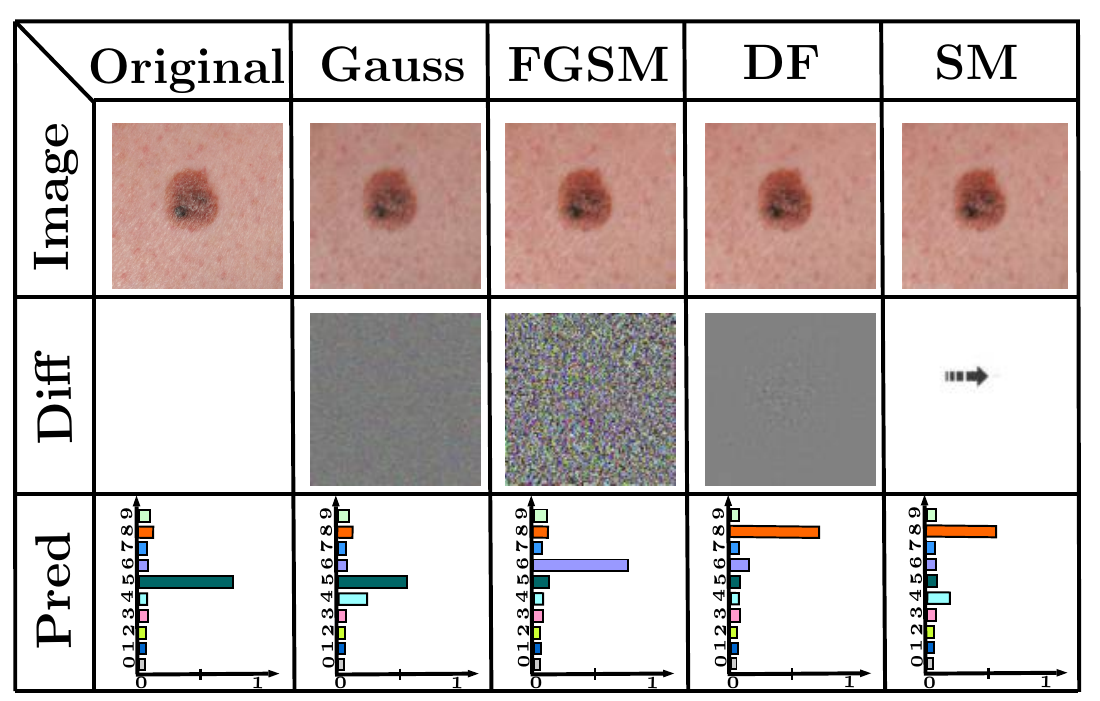}
(Left)
  \end{minipage}
  \hfill
  \begin{minipage}[t]{0.42\textwidth}
  \centering
\includegraphics[width=0.975\textwidth]{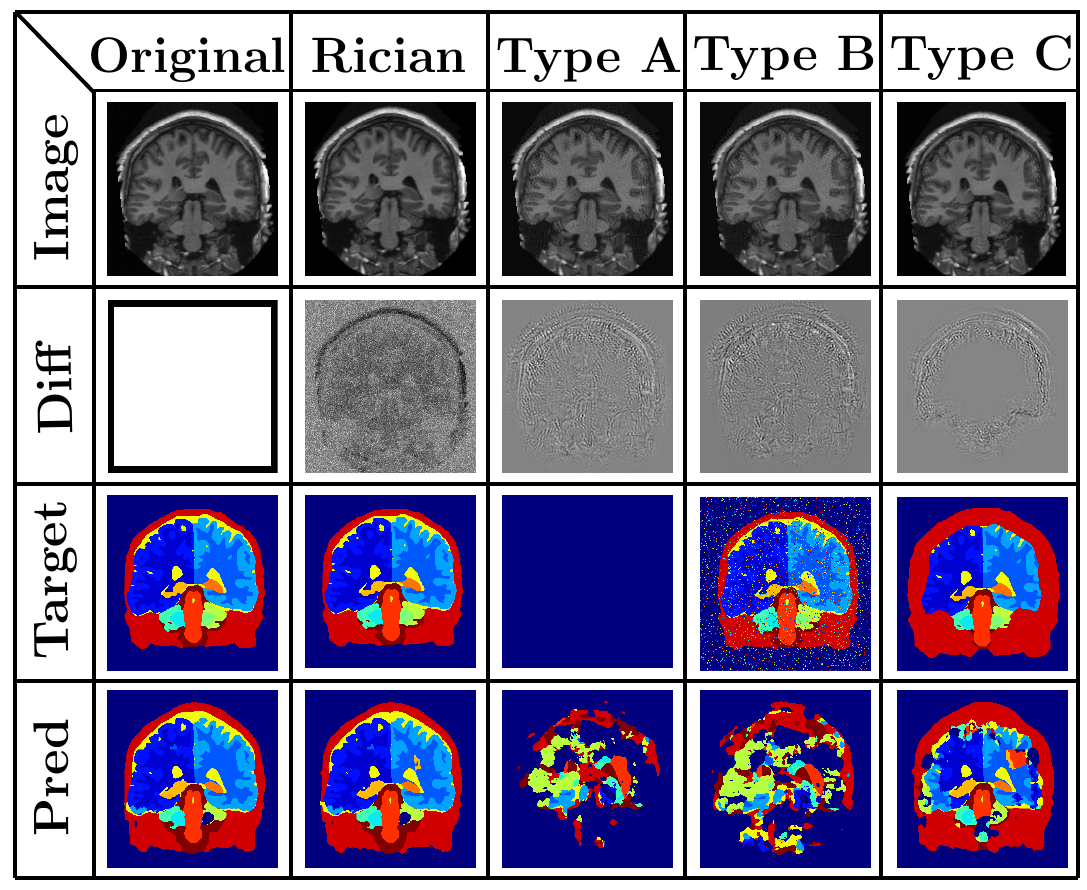}
(Right)
    \end{minipage}
  \end{minipage}

 \caption{Illustration of adversarial examples and their effect on model predictions. Left: Skin lesion classification and Right: Whole brain segmentation. Note that the added perturbation is effectively imperceptible to the human eye. Contrasting with prediction on original images, the crafted examples are able to successfully \textit{fool} the models into either misclassification or generating incorrect segmentation maps.}
  \label{fig:advEx}
\end{figure} 

\noindent
\textbf{Classification}: Gradient-based adversarial example generation methods have been proposed with the objective of generating minimum amount of perturbation $r$ that mis-classifies $\hat{X}$. These include the Fast Gradient Sign Method (FGSM)~\cite{goodfellow}, DeepFool (DF)~\cite{deepfool}, Saliency Map Attacks (SMA)~\cite{saliency_map_attack} \textit{etc.} Adversarial examples crafted with these methods are shown in Fig.~\ref{fig:advEx}.
For a trained model $F$, FGSM performs a one-step pixel-level update along the sign of the gradient that maximizes the task loss $J$ and the resultant perturbation is computed as $r = \epsilon\text{sign}\left ( \nabla_X J(\theta, X, Y) \right ) $, where $\theta$ are the parameters of the model. The amount of perturbation is regulated by a hyper-parameter $\epsilon$ that is typically assigned a low value, so that $\hat{X}$ is visually imperceptible from $X$. 

Differing from FGSM, DF follows an iterative greedy search process, where in each iteration the projections of the input sample to the decision boundaries of all the classes are computed and an $r$ is calculated that will push $X$ towards the closest decision boundary of a class, other than the correct one. In SMA, the impact of each pixel on the prediction of the model is estimated and the input is selectively perturbed to effect the most significant change to the output. 

\noindent
\textbf{Segmentation}: In~\cite{adversarial_segmentation}, the authors introduced Dense Adversarial Generation (DAG) as a method for crafting adversarial examples for semantic segmentation in a fast and effective way, closely resembling per-pixel, targeted FGSM. Particularly, DAG utilizes an incorrect segmentation mask, given by the user, and a target set of non-background pixels. Its goal is to calculate a minimum perturbation $r$ that will alter the prediction on the pixels in the target set from the correct class to the incorrect target class.


Specifically, DAG minimizes the distance between the prediction for $n$ pixels of the ground truth $Y = \{y_0, \dots, y_n\}$ and the incorrect target $Y'= \{y'_0, \dots, y'_n\}$ as shown in: $$L(X, T, Y, Y') = \sum_{n=1}^{N} [z_{y_n}(X, t_n) - z_{y'_n}(X, t_n)],$$ where $Z = \{z_0, \dots, z_{C-1}\}$ are the logits of the model and $C$ the number of classes. $T$ represents a target set of non-background pixels that DAG is allowed to perturb in order to constrain the search-space of the perturbation. In step $m$ the image has been transformed to $X_m = X + \sum_{m=0}^M r_m$, where the perturbation $r_m$ is computed by: $$r_m = \sum_{t_n \in T} [\nabla_{X_m} z_{y'_n}(X_m, t_n) - \nabla_{X_m} z_{y_n}(X_m, t_n)].$$

We utilized DAG to craft adversarial examples, seen in Fig.~\ref{fig:advEx}, by creating targets with varying degrees of difficulty. Particularly, we set the target to be all background (Type A), randomly assign a small percentage of pixels to a randomly-selected adversarial class (Type B) and modify (dilate) only a particular target class while keeping all other classes intact (Type C). Of the aforementioned attack types, Type A is the most challenging, causing the largest amount of perturbation, while Type C is expected to distort the image the least, as can be seen in Fig.~\ref{fig:advEx}. The Mean Square Error (MSE) between the original and adversarial images remained extremely small, ranging from 0.004 for adversaries of Type A to 0.002 for B and C.

\begin{figure}[t]
\centering
\begin{minipage}{\textwidth}
  \begin{minipage}[t]{0.63\textwidth}
\centering
\includegraphics[width=0.975\textwidth]{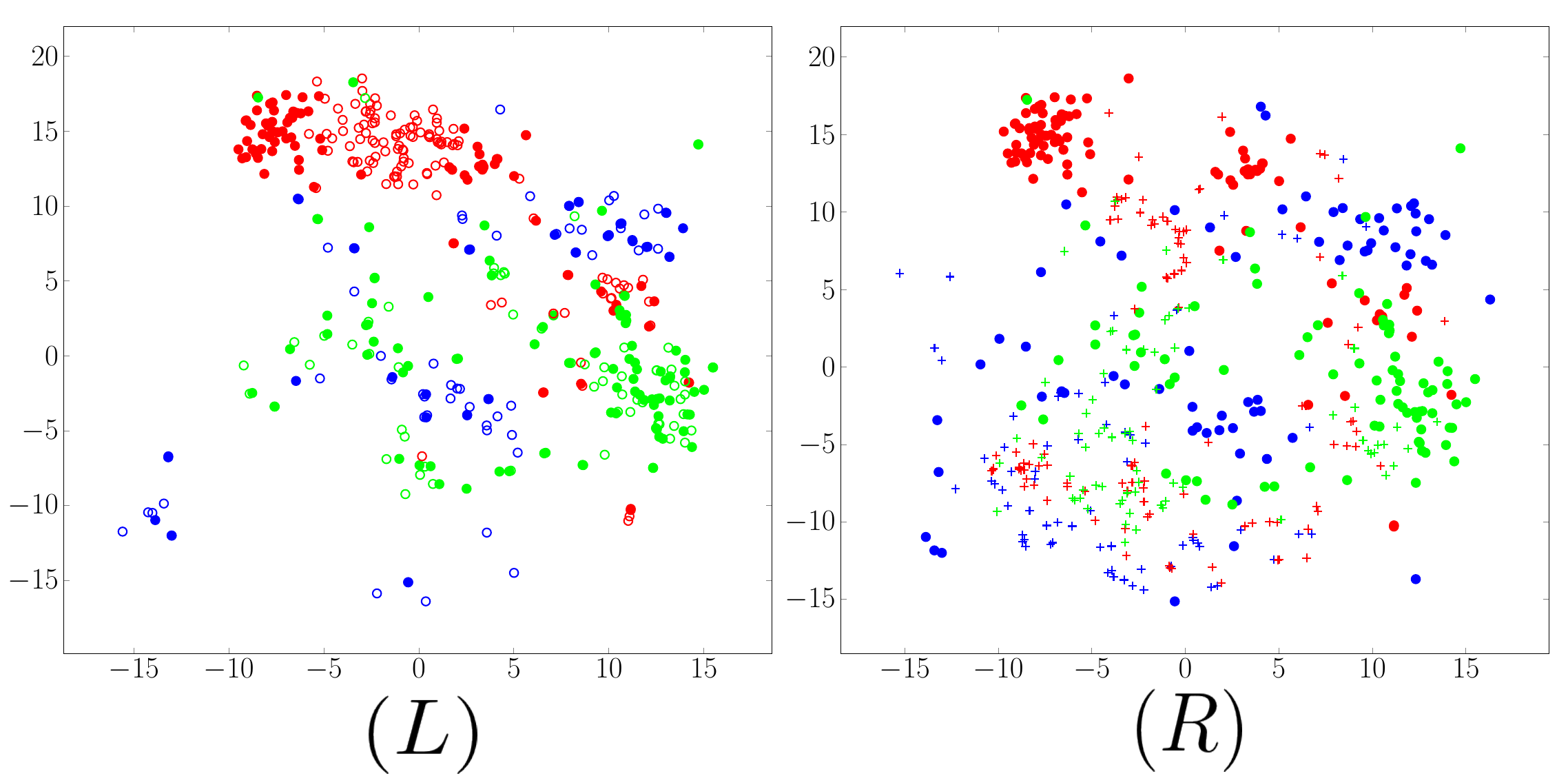}
  \end{minipage}
  \hfill
  \begin{minipage}[t]{0.36\textwidth}
  \vspace{-108pt}
\captionof{figure}{\small{t-SNE representation of the embeddings of 3 classes (red, blue and green) from clean ($\bullet $), noisy ($\circ$) and adversarial images ($+$). The noisy examples ($\circ$) are embedded closer to clean data (L), while adversarial ones are \textit{pushed} to the model boundaries (R).}}
  \label{fig:advTsne}
    \end{minipage}
  \end{minipage}
  \vspace{-0.7cm}
\end{figure}

\subsection{Model Evaluation With Adversarial Examples}
\label{ss:modLearn}

Our approach involves benchmarking models against a number of task-specific adversarial attacks discussed in Sec.~\ref{sec:advCraft} that are crafted implicitly from the data. The proposed pipeline for the evaluation of robustness is similar across both classification and segmentation. For classification, we crafted adversarial examples with FGSM, DF and SMA, while for segmentation we applied DAG with 3 different types of targets (Type A-C). 
Afterwards we attacked our models in a black-box fashion with examples generated by independently trained models, to maintain an unbiased attack scenario. In addition to that, black-box attacks highlight the \textit{transferability} of these examples between a variety of different models~\cite{black_box_attacks_transferability}.

\noindent
\textbf{Contrasting with Noise}: One could argue that applying noise on the test images before inference could replace the need for adversarial examples. However, that is not the case since hard ambiguous cases and outliers cannot be modeled by noise distributions. Adversarial examples, which are crafted with the purpose to force models to fail, are better suited for evaluating model behavior when subject to input extrema. To showcase that adversarial perturbations do not resemble noise distributions, we also crafted images distorted with modality-specific noise (Gaussian noise for dermatoscopic images and Rician noise for T1w MRI). 
For fairness, the Structural Similarity (SSIM) between the original and noisy images was the same as the one between the original and adversarial examples and ranged from 0.97 to 0.99. 

We plot the t-Stochastic Neighbor Embedding representation (t-SNE) from IV3 for the clean, noisy and adversarial examples (FGSM) in Fig.~\ref{fig:advTsne} for the classification task to further illustrate this difference. Contrasting Fig.~\ref{fig:advTsne}(L) with Fig.~\ref{fig:advTsne}(R), we clearly observe that images distorted with noise are embedded close to the clean images, while adversarial examples are pushed further towards other classes. The anomalous nature of the adversarial examples clearly supports our hypothesis that their behavior is not akin to noise and can act as a harder benchmark for evaluating a model's robustness.

\section{Experiments}

To provide a proof-of-concept for the proposed robustness evaluation we chose the challenging tasks of fine-grained skin lesion classification using dermatoscopic images and segmentation of the whole brain in T1w-MRI scans. The task-specific model learning is described as follows:  

\noindent
\textbf{Classification}: We fine-tune three state-of-the art deep learning architectures namely, InceptionV3 (IV3)~\cite{inception}, InceptionV4 (IV4)~\cite{inception} and MobileNet (MN)~\cite{mobilenet} for this task. Both IV3 and IV4 are very-deep architectures ($>$ 100 layers), while MN is significantly compact. Comparing these architectures would help discover if any innate relationships exist between model complexity (in terms of depth and parameters) and their robustness. To keep the comparisons fair, all the models were initialized with their respective ImageNet parameters and fine-tuned with a weighted cross-entropy loss with affine data augmentation. Specifically, the models were trained with stochastic gradient descent with a decaying learning rate initialized at 0.01, momentum of 0.9 and dropout of 0.8 for regularization. We use the publicly-available Dermofit~\cite{dermofit} image library consisting of 1300 high-quality dermatoscopic images, with histologically validated fine grained expert annotations (10 classes) for this task. The dataset was split at patient-level with non-overlapping folds (50\% for training and rest for testing).

\noindent
\textbf{Segmentation}: For this task we chose to evaluate three popular fully-con\-vo\-lu\-ti\-o\-nal deep architectures, namely SegNet (SN)~\cite{segnet}, UNet (UN)~\cite{unet} and DenseNet (DN)~\cite{densenet}. Contrasting across these architectures, we evaluate the importance of skip connections with respect to robustness varying from no skip connections in SN to introducing long-range skips in UN and both long and short-range skip connections in DN. The model parameters (depth and layers) were chosen to maintain comparable model complexity, so as to exclusively factor out the impact of skip connections to robustness. The aforementioned models were trained with a composite loss of weighted-cross entropy and Dice loss as proposed in~\cite{error_corrective_boosting} and model optimization was performed with ADAM optimizer with an initial learning rate of 0.001. We use 27 volumes from the publicly-available whole-brain segmentation benchmark (subset of Open Access Series of Imaging Studies (OASIS) dataset~\cite{oasis}) that was released as a part of the Multi-Atlas Labeling Challenge in MICCAI 2012~\cite{malc}, with 80-20 patient-level splits for training and testing. 
The models for both tasks were trained until convergence using the TensorFlow~\cite{tensorflow} deep learning library and adversarial examples for DF and SMA attacks described in Sec.~\ref{ss:modLearn} were crafted using the FoolBox~\cite{foolbox} library. 
\\ \indent Following the model evaluation strategy presented in Sec.~\ref{ss:modLearn}, adversarial examples were crafted for each of the trained models and their robustness is evaluated in terms of average classification accuracy and average Dice score. We report the overall performance of the models in Table~\ref{tab:all_results}, where we compare the performance of each model on clean and noisy images with their average score against all the attacks. Furthermore, in Table~\ref{tab:black_box_results} we are reporting the performance of each model against all the black-box attacks separately. 

\begin{figure}[t]
\centering
\begin{minipage}{\textwidth}
  \begin{minipage}[t]{0.45\textwidth}
\centering
\includegraphics[width=0.975\textwidth]{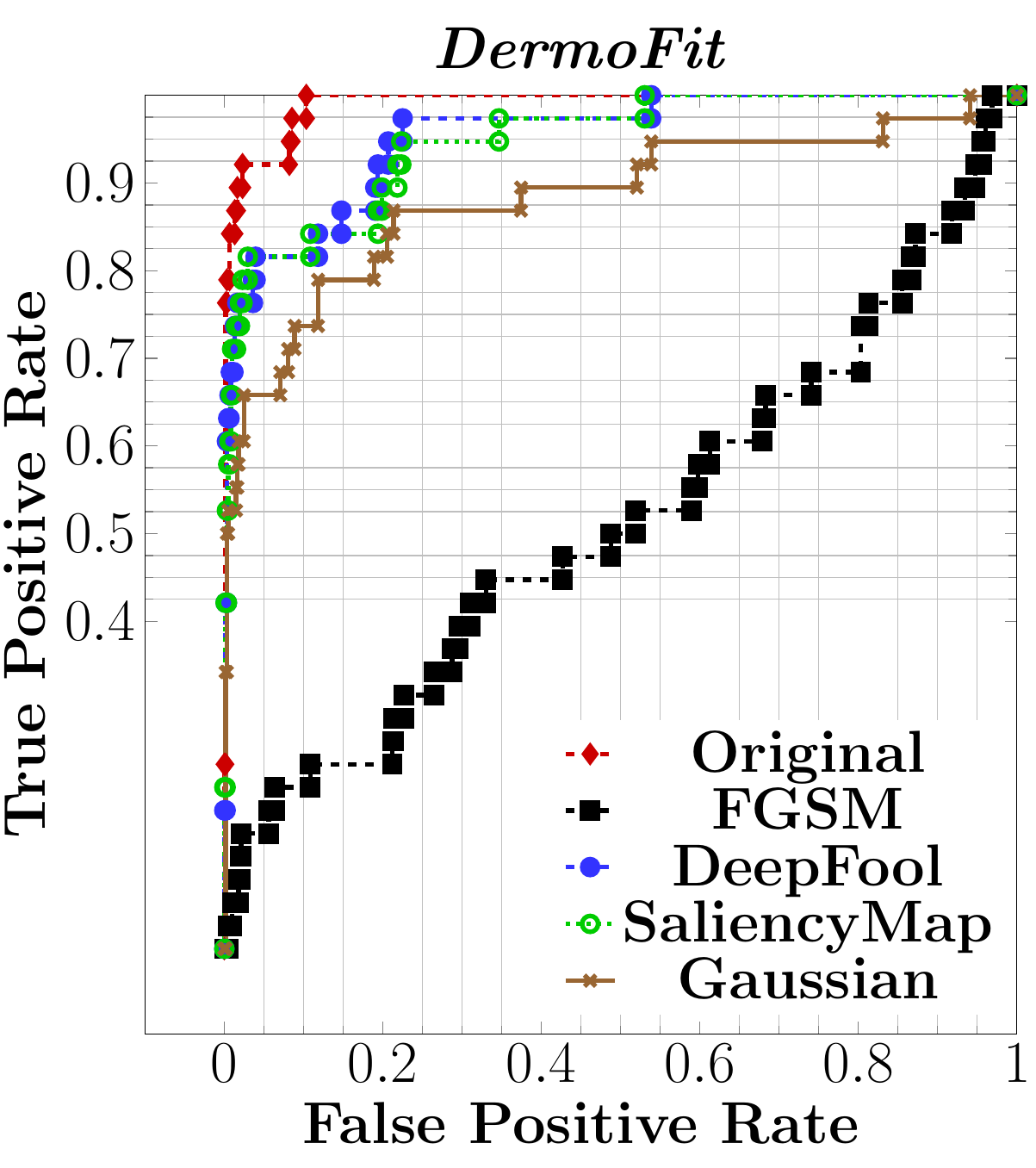}
  \end{minipage}
  \hfill
  \begin{minipage}[t]{0.5\textwidth}
  \centering
\includegraphics[width=0.975\textwidth]{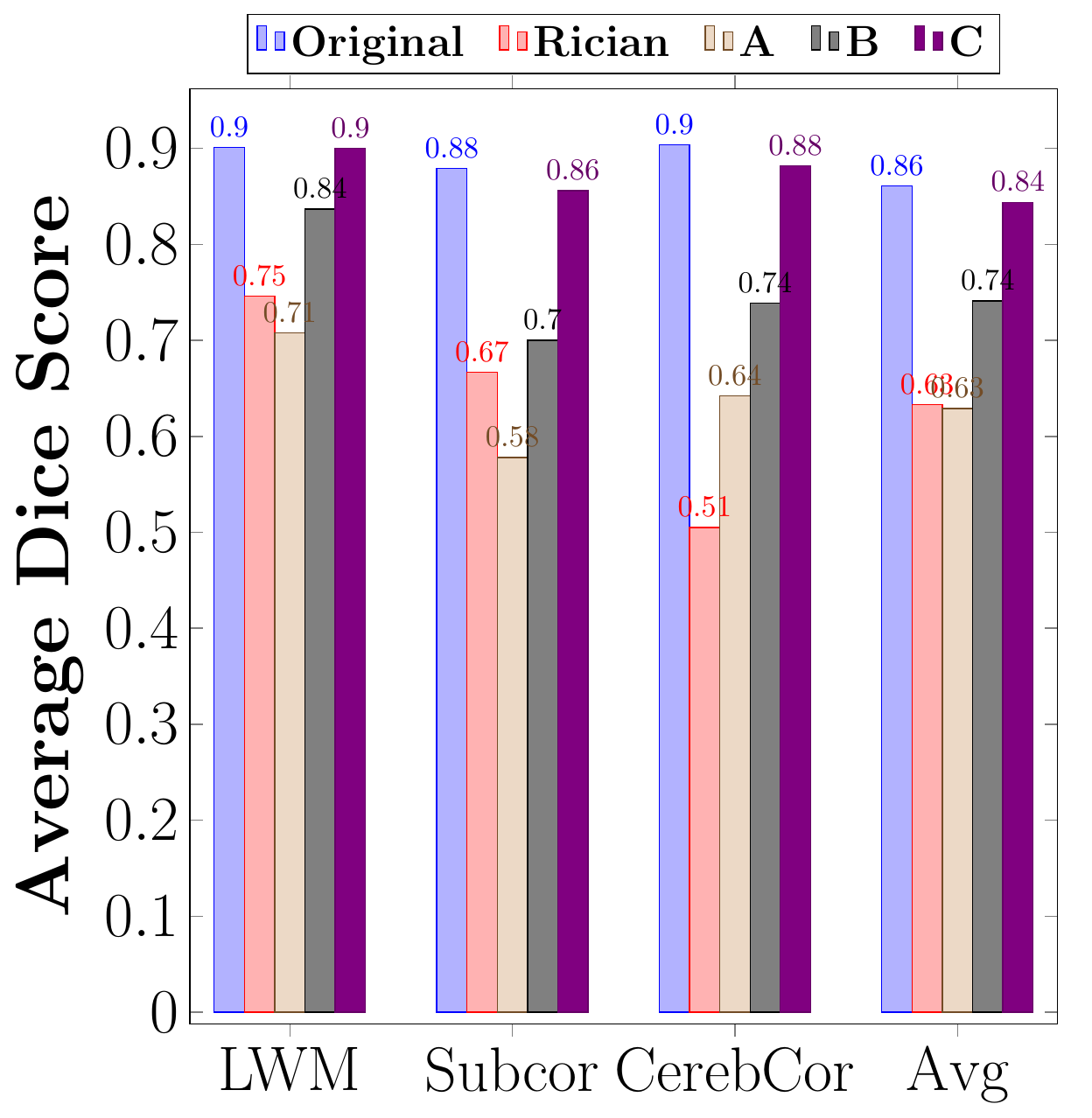}
    \end{minipage}
  \end{minipage}

 \caption{(Left) ROC Curves for InceptionV3 on clean, noisy and adversarial images. Notice the varying drop in performance for the different types of attacks. (Right) Bar Chart for DenseNet showing the Dice Score achieved for 3 structures as well as the average across all structures against 3 types of attacks and images distorted with Rician noise.}
  \label{fig:ROC_BAR}
\end{figure}

We also report the ROC curves for the class of malignant melanoma in Fig.~\ref{fig:ROC_BAR} (Left) and the average and structure-wise Dice score for prominent structures in Fig.~\ref{fig:ROC_BAR} (Right).

\section{Results and Discussion}

\subsection{Robustness Evaluation for Classification}
\begin{figure}[t]
\centering
\begin{minipage}{\textwidth}
  \begin{minipage}[t]{0.55\textwidth}
\centering
\begin{tabular}{|l|l|c|c|c|c|}

\multicolumn{3}{c}{\textbf{ }} &   \textbf{Noise} &            \multicolumn{2}{c}{\textbf{Adversarial}}   \\ \hline 
\multirow{4}{*}{\rotatebox[origin=c]{90}{\tiny{Classification}}} &             & \textbf{Clean} & \textbf{Gaussian} & \textbf{Avg}   & \textbf{\% Drop} \\ \cline{2-6} 
                   & \textbf{IV3}~\cite{inception} & 0.710          & 0.693             & \textbf{0.641} & \textbf{6.897}            \\ \cline{2-6} 
                   & \textbf{IV4}~\cite{inception} & \textbf{0.810} & \textbf{0.761}    & 0.633          & 17.72   \\ \cline{2-6} 
                   & \textbf{MN}~\cite{mobilenet} & 0.800          & 0.647             & 0.564          & 24.55            \\ \cline{1-6} 
\multirow{4}{*}{\rotatebox[origin=c]{90}{\tiny{Segmentation}}} &             & \textbf{Clean} & \textbf{Rician}   & \textbf{Avg}   & \textbf{\% Drop} \\ \cline{2-6} 
                   & \textbf{SN}~\cite{segnet}  & 0.842          & 0.595             & 0.470          & 37.17            \\ \cline{2-6} 
                   & \textbf{UN}~\cite{unet}  & \textbf{0.862} & 0.759             & 0.453          & 40.92   \\ \cline{2-6} 
                   & \textbf{DN}~\cite{densenet}  & 0.861          & \textbf{0.848}    & \textbf{0.667} & \textbf{19.53}            \\ \cline{1-6} 
\end{tabular}
\label{tab:all_results}
  \end{minipage}
  \hfill
  \begin{minipage}[t]{0.42\textwidth}
  \vspace{-1.9cm}
 \captionof{table}{Comparative evaluation of the classification and segmentation models on clean, noisy and adversarial examples. We report the average accuracy and Dice overlap score along with the \% drop in performance on adversarial examples with respect to performance on clean data.}
  \label{tab:all_results}
    \end{minipage}
  \end{minipage}
\vspace{-0.2cm}
\end{figure} 
\noindent
\textbf{Visual Evaluation}: Fig.~\ref{fig:advEx} (Left) illustrates adversarial examples crafted for an unseen test example (belonging to malignant melanoma class) for each of the classification related attacks (FGSM, DF and 
SMA) alongside an image perturbed with Gaussian noise for comparison. A scaled version of the difference image with respect to the original is also shown along with the posterior probabilities estimated using IV3 network. We observe that all the adversarial examples, regardless of the attack are consistently misclassified with very high confidence, while the addition of Gaussian noise only results in confidence reduction. Furthermore, FGSM induces perturbations dispersed across the whole image, while DF and SMA generate perturbations more localized to the lesion.   



\noindent
\textbf{Attacks}: From Table~\ref{tab:all_results}, we observe that IV4 and MN both achieve comparable performance on clean data (80-81\%) superior to the one of IV3 (71\%). By limiting model evaluation to generalizability (\textit{i.e.} performance on clean data) one may prematurely conclude that IV3 demonstrates the worst comparative performance. However, upon comparing the robustness of these models with respect to average performance under all the attacks (Table~\ref{tab:black_box_results}), we observe a contrary trend. The performance drop for IV3 is significantly lower (7\%) in comparison to IV4 (17\%) and MN (25\%). IV4 achieves higher accuracy on DF and SMA attacks, while IV3 is the most robust model against FGSM. Contrasting IV4 and MN, we observe that MN performs poorly not only on noisy samples but also on all of the attacks, as shown in Table~\ref{tab:black_box_results}. These contrasting observations clearly substantiate the core hypothesis within the paper that model evaluation should not be limited to generalizability and that performing robustness evaluation is equally important. Despite showing comparable performance in terms of generalizability, we can clearly conclude that IV4 is strongly preferred over MN.

\begin{table}[t]
\centering
\resizebox{\textwidth}{!}{
\begin{tabular}{|l|c|c|c|c|c|c|c|c|c|c|}
\cline{1-11}
\multirow{5}{*}{\rotatebox[origin=c]{90}{\tiny{Classification}}} &  & \multicolumn{3}{c|}{\textbf{FGSM}}               & \multicolumn{3}{c|}{\textbf{DF}}                 & \multicolumn{3}{c|}{\textbf{SMA}}               \\ \cline{2-11}
                   & \textbf{}   & \textbf{IV3}    & \textbf{IV4}    & \textbf{MN}    & \textbf{IV3}    & \textbf{IV4}    & \textbf{MN}    & \textbf{IV3}    & \textbf{V4}    & \textbf{MN}                \\ \cline{2-11} 
                   & \textbf{IV3}~\cite{inception} & \textbf{0.449} & \textbf{0.548} & \textbf{0.567} & 0.729          & 0.707          & 0.664          & \textbf{0.738} & 0.701          & 0.669                       \\ \cline{2-11} 
                   & \textbf{IV4}~\cite{inception} & 0.429          &          0.411      & 0.451          & \textbf{0.743} &         \textbf{0.768}       & \textbf{0.697} & 0.735          &      \textbf{0.778}          & \textbf{0.683}                              \\ \cline{2-11} 
                   & \textbf{MN}~\cite{mobilenet} & 0.335          & 0.275          & 0.213          & 0.726          & 0.731 & 0.672          & 0.732          & 0.735 & 0.661                                     \\ \cline{1-11} 
\multirow{5}{*}{\rotatebox[origin=c]{90}{\tiny{Segmentation}}} & \textbf{}   & \multicolumn{3}{c|}{\textbf{Type A}}             & \multicolumn{3}{c|}{\textbf{Type B}}             & \multicolumn{3}{c|}{\textbf{Type C}}            \\ \cline{2-11}
                   & \textbf{}   & \textbf{SN}     & \textbf{UN}     & \textbf{DN}     & \textbf{SN}     & \textbf{UN}     & \textbf{DN}     & \textbf{SN}     & \textbf{UN}     & \textbf{DN}                                    \\ \cline{2-11} 
                   & \textbf{SN}~\cite{segnet}  & 0.277          & 0.272          & 0.309          & 0.397          & 0.473          & 0.428          & 0.669          & 0.702          & 0.705                                     \\ \cline{2-11} 
                   & \textbf{UN}~\cite{unet}  & 0.248          & 0.434          & 0.258          & 0.364          & 0.434          & 0.368          & 0.636          & 0.653          & 0.677                                    \\ \cline{2-11} 
& \textbf{DN}~\cite{densenet}  & \textbf{0.600} & \textbf{0.528} & \textbf{0.415} & \textbf{0.749} & \textbf{0.721} & \textbf{0.563} & \textbf{0.819} & \textbf{0.791} & \textbf{0.814}                      \\ \cline{1-11} 
\end{tabular}
}
\vspace{0.2cm}
\caption{Comparative evaluation of model robustness using black-box attacks for the tasks of classification and segmentation. We report the average accuracy for classification and average Dice overlap score across structures for segmentation.}
\vspace{-0.2cm}
\label{tab:black_box_results}
\end{table}

\subsection{Robustness Evaluation for Segmentation}
\noindent
\textbf{Visual Evaluation}: Fig.~\ref{fig:advEx} (Right) illustrates how the prediction maps of the trained DN model transform when it is attacked by adversarial examples. All the DAG attacks (Type A-C) successfully fool the model into producing an incorrect prediction map. However, the prediction on the image distorted with Rician noise is visually very similar to the one of the original image and the ground truth. This clearly demonstrates that adding adversarial perturbation is not akin to adding randomly generated noise. Thus, adversarial examples are superior in pushing a network to its limits and evaluating its robustness. 

\noindent
\textbf{Attacks}:  From Table~\ref{tab:all_results}, we observe that DN (86.1\%) and UN (86.2\%) achieve almost identical performance on clean unseen test examples and fare better than SN (84.2\%), highlighting the importance of skip connections. Furthermore, the fact that the performance drop caused by the addition of Rician noise remains low for UN and DN (10\% and 1\% respectively) reinforces the distinction between noise and adversarial perturbations.
Regarding model performance with respect to adversarial attacks in Table~\ref{tab:black_box_results}, we observe that DN is not only resilient to noise but also significantly more robust than SN (by 18\%) and UN (by 21\%) to attacks crafted by any other model. Furthermore, SN and UN remain highly vulnerable to all the attacks with a significant 37-40\% drop in their average Dice score.
In consensus to the classification results discussed earlier, comparing the three segmentation models only in terms of generalizability would not have been sufficient to determine the best one. Both its resilience to samples distorted with Rician noise and its consistent resistance to adversarial attacks make DN the strongest model among its competitors for this task.

\section{Conclusion}
In this paper, for the first time, we explored adversarial examples in medical imaging for the tasks of classification and segmentation. 
We propose a strategy for model evaluation by leveraging task-specific adversarial attacks, that evaluate not only a model's generalizability, but also its robustness. We showed that for two models with comparable performance, their relative exploration of the underlying data manifold may have significant differences, hence resulting in varying robustness and model sensitivities. Specifically, we demonstrate that for segmentation tasks the use of dense blocks and skip connections contributes to both improved generalizability and robustness, while model depth seems to increase the resistance of classification models to adversarial examples.

\end{document}